\documentclass[journal,twoside,web]{ieeecolor}
\usepackage[T1]{fontenc}
\usepackage{generic}
\definecolor{subsectioncolor}{RGB}{0,0,0}
\usepackage{cite}
\usepackage{amsmath,amssymb,amsfonts}
\usepackage{algorithm,algorithmic}
\usepackage{graphicx}
\usepackage{booktabs}
\usepackage{array}
\newcolumntype{L}[1]{>{\raggedright\arraybackslash}p{#1}}
\usepackage{url}
\usepackage{hyperref}
\hypersetup{hidelinks,pdfauthor={Hien M. Tran, Hien T. T. Pham, Tan D. Le, Hieu X. Nguyen, Anh P. T. Mai}}
\usepackage{textcomp}
\graphicspath{{figures/}}
\makeatletter
\def\revisionhead#1#2{\rlap{\raisebox{-5pt}[0pt][0pt]{\color{subsectioncolor}\rule{\textwidth}{0.5pt}}}\hbox to\textwidth{\sffamily\scriptsize #1\hfil #2}}
\def\ps@headings{\def\@oddhead{\revisionhead{\rightmark}{\thepage}}\def\@evenhead{\revisionhead{\thepage}{\leftmark}}\let\@oddfoot\@empty\let\@evenfoot\@empty}
\def\ps@titlepagestyle{\def\@oddhead{\revisionhead{\leftmark}{\thepage}}\def\@evenhead{\revisionhead{\thepage}{\leftmark}}\let\@oddfoot\@empty\let\@evenfoot\@empty}
\makeatother
\usepackage{color}

\begin{document}
\bstctlcite{IEEEexample:BSTcontrol}
\title{An Automated Framework Integrating Deep Learning and Anatomical Reasoning for Cardiac Phenotyping From CCTA}
\author{Hien~Minh~Tran, Thi-Thu-Hien~Pham, Le~Duy~Tan, Hieu~Xuan~Nguyen, and Tuong-Anh~Mai-Phan
\thanks{Hien~Minh~Tran, Le~Duy~Tan, Hieu~Xuan~Nguyen, and Tuong-Anh~Mai-Phan are with the University of Health Sciences, Vietnam National University Ho Chi Minh City, Ho Chi Minh City, Vietnam. Thi-Thu-Hien~Pham is with the Ho Chi Minh City University of Technology, Ho Chi Minh City, Vietnam.}
\thanks{\raggedright E-mail: \mbox{Drtranminhhien@gmail.com} ; \mbox{phamthithuhien@iuh.edu.vn}; \mbox{ldtan@hcmiu.edu.vn} ; \mbox{nguyen.hieuee1810@gmail.com}; \mbox{mptanh@uhsvnu.edu.vn}.}
\thanks{\raggedright Corresponding author: Tuong-Anh~Mai-Phan (e-mail: \mbox{mptanh@uhsvnu.edu.vn}).}\endgraf}
\maketitle

\begin{abstract}
Coronary computed tomography angiography (CCTA) depicts coronary anatomy and cardiac fat, but segmentation alone does not ensure anatomically consistent measurements. We propose an automated framework that combines deep learning with anatomical reasoning for quantitative cardiac phenotyping. Its anatomical measurement module uses graph geometry, branch identity, and surrounding structures to define vessel-specific pericoronary adipose tissue (PCAT) sampling and a surrogate epicardial adipose tissue (EAT) compartment. The latter also generates training targets without requiring expert pericardium contours. A companion module quantifies coronary calcium on non-contrast CT. Component-wise benchmarking was conducted on five public datasets: ImageCAS, ImageCAS-X, COCA, Visual Lab, and MultiD4CAD. On ImageCAS-X Test-160, lumen segmentation with validation-selected post-processing achieved a Dice score of 0.9133 and a clDice score of 0.940, exceeding the corresponding published CAS-Net point estimates. In the four-split ImageCAS evaluation, the four-member ensemble outperformed the single ResEnc-L model. Analysis of 800 paired scans showed that PCAT attenuation measurements differed between whole-vessel and lumen references. External EAT evaluation showed strong volume correlation and good spatial overlap. Explicit anatomical rules make the resulting measurements interpretable and inspectable. In practice, the framework provides a standardized pipeline for processing cardiac CT cohorts and quantifying coronary calcium, PCAT, and EAT for population-level analysis and comparison across studies. The source code is publicly available at \url{https://github.com/HarryTRAN207/CardiacCT-Research}.
\end{abstract}

\begin{figure*}[!t]
\centering
\includegraphics[width=\textwidth]{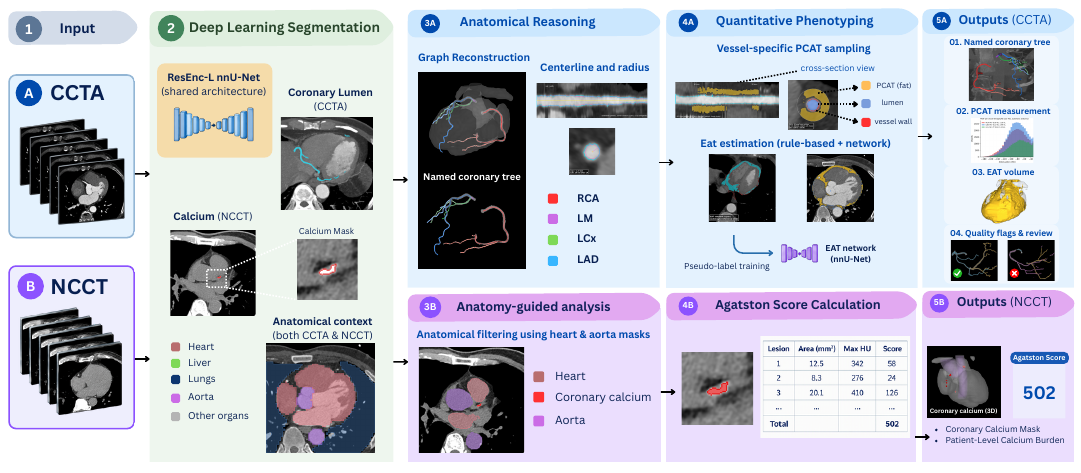}
\caption{Automated cardiac phenotyping from CCTA and companion calcium scoring from NCCT: (1) inputs; (2) separately trained ResEnc-L nnU-Net models and TotalSegmentator context; (3A) anatomical reasoning to a named coronary tree and (3B) anatomy-guided calcium filtering; (4A) vessel-specific PCAT and rule-based or learned EAT, (4B) Agatston scoring; (5) outputs with quality flags.}
\label{fig:pipeline}
\end{figure*}

\section{Introduction}
\label{sec:intro}

\IEEEPARstart{C}{oronary} CT angiography (CCTA) supports assessment of coronary anatomy and additional imaging markers of cardiovascular risk~\cite{lin2022lancetdh}. Pericoronary adipose tissue (PCAT) attenuation reflects vascular inflammation, and the fat attenuation index (FAI) predicted cardiac mortality in the CRISP-CT cohorts~\cite{antonopoulos2017,oikonomou2018,oikonomou2019ml}. Epicardial adipose tissue (EAT) volume is associated with incident coronary events independently of conventional risk factors~\cite{mahabadi2013,iacobellis2022}.

Three challenges motivate this work. First, overlap accuracy does not establish anatomical usability: disconnected vessels, ambiguous bifurcations and incorrect branch names can disrupt proximal sampling despite high Dice. Phenotyping must therefore connect voxel predictions to topology, vessel identity and measurement regions.

Second, whole-vessel and lumen references define different boundaries and PCAT sampling regions~\cite{zeng2023imagecas,imagecasx2026}. Measurement differences can reflect both reference choice and segmentation error. EAT also requires a pericardial compartment, yet expert contours are not consistently available~\cite{commandeur2018}. Anatomical constraints can reduce this annotation requirement but require independent evaluation.

Third, public datasets provide complementary rather than complete phenotype references. Shared scans require model-specific partitions~\cite{kapoor2023leakage}, while acquisition settings and reference definitions complicate external comparisons~\cite{ma2020fai}. Evaluation must connect component accuracy to measurement agreement and retain incomplete outputs.

Our automated workflow combines nnU-Net coronary segmentation~\cite{isensee2021,isensee2024revisited}, TotalSegmentator context~\cite{wasserthal2023} and anatomical reasoning (Fig.~\ref{fig:pipeline}). Tree geometry and branch identity define PCAT sampling; the cardiac frame and left-main landmark constrain an EAT surrogate and training targets. This shared representation connects named trees, PCAT, EAT and quality flags, with companion non-contrast calcium scoring.

Paired anatomy-source analysis and external testing characterize reference and acquisition dependencies. The main contributions of this work are as follows:
\begin{enumerate}
\item An anatomical measurement construction method coupling coronary graphs, branch identity and cardiac context to PCAT/EAT quantification and rule-derived EAT supervision without expert pericardium contours.
\item Competitive coronary segmentation with ImageCAS-X lumen Dice 0.9133 and clDice 0.940, and ImageCAS Dice of 0.8416 and 0.8518 across four official splits for the single model and four-member ensemble, respectively.
\item Benchmarking on five public datasets, with paired PCAT reference effects of 3.0--4.2~HU and external EAT agreement ($r=0.944$; matched Dice 0.800).
\end{enumerate}

\section{Related Work}
\label{sec:related}

\subsection{Deep Learning for Coronary Segmentation}
ImageCAS~\cite{zeng2023imagecas} supports coronary-specific feature fusion in 3D-FFR-UNet~\cite{song2022ffr}, refinement in Seg2RefineNet~\cite{seg2refinenet2025}, diffusion attention in DiffCAS~\cite{diffcas2024} and small-vessel recovery in Ensembled-SAMs~\cite{chen2025ensembledsams}. CAS-Net uses multi-scale attention~\cite{dong2023casnet}; ADE-HTL constrains anatomy and topology~\cite{zhang2024adehtl}. Transformer, adaptive-token and uncertainty-aware designs further model context and boundaries~\cite{hatamizadeh2022swinunetr,du2025tdformer,wang2026dpgnet}. These methods still require branch identification and sampling-region construction for phenotyping. ImageCAS-X distinguishes lumen from whole-vessel references~\cite{imagecasx2026}, affecting segmentation scores and downstream PCAT.

\subsection{Coronary Anatomy and Perivascular Quantification}
Centreline benchmarks and trackers extract coronary geometry~\cite{schaap2009,wolterink2019centerline}; reporting conventions and tree models establish branch identities~\cite{austen1975,leipsic2014scct,wu2019labeling}. Nguyen et al.~\cite{nguyen2024pcat} automated RCA/LCA PCAT using segmentation, anatomical context and centrelines, but did not report paired agreement against expert coronary anatomy or EAT assessment. Although perivascular fat has biological and prognostic relevance~\cite{antonopoulos2017,oikonomou2018}, acquisition effects~\cite{ma2020fai} and sampling definitions require workflow-specific validation.

\subsection{Cardiac Adipose Tissue and Calcium Assessment}
Supervised EAT methods use expert anatomical annotations~\cite{commandeur2018}; image-level supervision has reduced annotation requirements in optical coherence tomography~\cite{huang2023cat}. Defining the pericardial compartment without expert contours remains a CT challenge. Visual Lab and MultiD4CAD offer complementary references with different acquisition and annotation definitions~\cite{rodrigues2016,prinzi2025multid4cad}.

Automated Agatston scoring~\cite{agatston1990} spans cardiac and chest CT~\cite{wolterink2016,lessmann2018,vanvelzen2020,eng2021}. Anatomical priors constrain calcium segmentation~\cite{alirr2025}, but patient-level scoring requires separate evaluation, and contrast-enhanced CCTA requires dedicated calibration~\cite{mylonas2012}.

\subsection{Integrated Workflows and Study Positioning}
Integrated workflows combine complementary processing and learning stages across biomedical applications, including ECG-based sleep-apnea classification~\cite{nguyen2024mpcnn} and instruction-driven medical image segmentation~\cite{huang2026medsegagent}. Cardiac phenotyping additionally requires anatomical definitions, measurement validation~\cite{xue2021lvquan,maierhein2024metrics} and leakage checks~\cite{kapoor2023leakage}. Our framework connects coronary reconstruction, branch naming and PCAT/EAT measurement through anatomical reasoning, generates EAT targets without expert pericardium contours, and evaluates anatomy-source and external measurement dependencies.

\section{Methods}
\label{sec:methods}

\subsection{Problem Definition}
Let $X:\Omega\to\mathbb{R}$ be a CCTA volume in HU, with voxel centres mapped to physical coordinates by the image affine. Anatomical masks are voxel sets; spacing $(s_x,s_y,s_z)$ gives voxel volume $\Delta v=s_xs_ys_z$ in mm$^3$. Define
\begin{equation}
\begin{aligned}
d(\mathbf{x},S)&=\min_{\mathbf{y}\in S}\|\mathbf{x}-\mathbf{y}\|_2,\\
\operatorname{Vol}(M)&=|M|\Delta v/1000\quad\text{mL},
\end{aligned}
\label{eq:basic}
\end{equation}
where distances are in mm and $|M|$ counts voxels.
Given $X$, the framework returns coronary mask $M_C$, named trees $\mathcal{T}$, PCAT measurements for $\mathcal{V}=\{\mathrm{RCA},\mathrm{LAD},\mathrm{LCx}\}$, EAT mask $M_E$ and volume, and quality flags $\mathbf{q}$ (Fig.~\ref{fig:pipeline}). LM denotes the common LAD/LCx trunk. An optional non-contrast volume $X_{\mathrm{NC}}$ yields calcium mask $K$ and Agatston score $A$.

\subsection{Deep Learning and Anatomical Reasoning}
\label{sec:seg}

The framework first estimates coronary and surrounding
anatomy:
\begin{equation}
p=f_{\theta}(X),
\qquad
\mathcal{A}=g(X),
\label{eq:anatomy}
\end{equation}
where $p$ is the coronary probability map predicted by
nnU-Net v2 with the ResEnc-L
preset~\cite{isensee2021,isensee2024revisited}, and
$\mathcal{A}$ is the anatomical context obtained from
TotalSegmentator~\cite{wasserthal2023}.
It includes the heart $H$, aorta $A_o$, and surrounding
structures used for anatomical constraints.

Whole-vessel, lumen and calcium predictors are trained separately for their respective targets; sharing the ResEnc-L architecture does not imply shared weights.
The selected model and its post-processing configuration
determine $M_C$.

\subsubsection{Coronary post-processing}

The post-processing configuration selected on validation-80
and evaluated on test-160 uses hysteresis with a lower threshold $\tau_l=0.3$ and an upper seed
threshold $\tau_h=0.5$. The M1 baseline uses
threshold 0.5 without component filtering. Define
\[
\begin{aligned}
L_{\mathrm{low}}&=\{\mathbf{x}:p(\mathbf{x})\geq\tau_l\},\\
L_{\mathrm{seed}}&=\{\mathbf{x}:p(\mathbf{x})\geq\tau_h\}.
\end{aligned}
\]
Let $\operatorname{CC}_{3D}(M)$ denote the connected
components of a three-dimensional mask $M$.
A component is retained when it contains a seed and has
a volume of at least 10~mm$^3$:
\begin{equation}
\begin{aligned}
\mathcal{C}_{\mathrm{keep}}&=\{C\in\operatorname{CC}_{3D}(L_{\mathrm{low}}):
 b(C)=1\},\\
b(C)&=\mathbf{1}[C\cap L_{\mathrm{seed}}\neq\varnothing]\,
\mathbf{1}[|C|\Delta v\geq10~\mathrm{mm}^3],\\
M_C&=\bigcup_{C\in\mathcal{C}_{\mathrm{keep}}}C.
\end{aligned}
\label{eq:postprocess}
\end{equation}
Thus, lower-confidence voxels are retained only when
connected to a high-confidence region, and small isolated
components are removed.

\subsubsection{Anatomical measurement construction}

\begin{figure*}[!t]
\centering
\includegraphics[width=\textwidth]{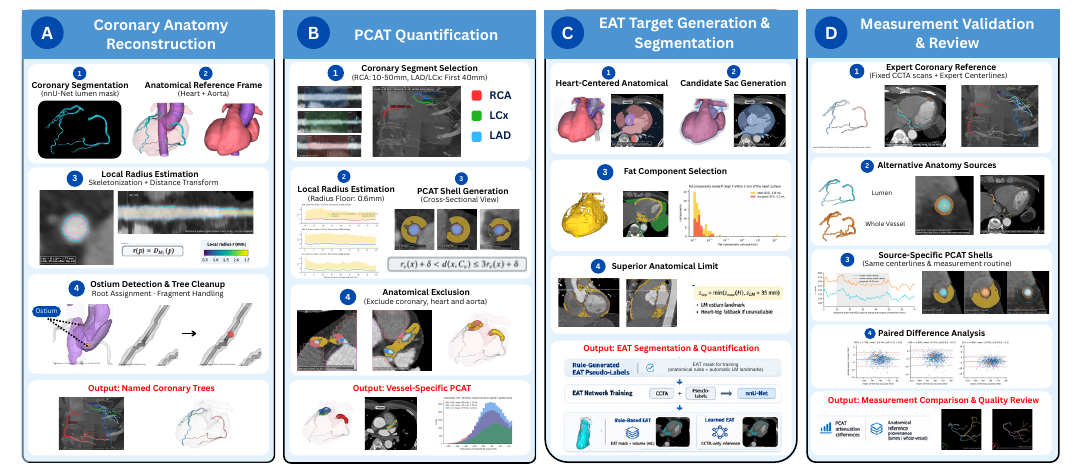}
\caption{Anatomical reasoning for cardiac phenotyping. (A) Coronary anatomy reconstruction. (B) Vessel-specific PCAT quantification. (C) Anatomically constrained EAT target generation and segmentation. (D) Paired anatomy-source comparison and measurement review.}
\label{fig:anatomical_reasoning}
\end{figure*}

Algorithm~\ref{alg:pipeline} summarizes phenotype extraction.
Figure~\ref{fig:anatomical_reasoning} connects four stages: coronary reconstruction (A), vessel-specific PCAT sampling (B), EAT target generation and segmentation (C), and paired measurement validation and review (D). Training and model-specific configurations are given in Section~\ref{sec:setup}.

\begin{algorithm}[!t]
\caption{Automated phenotype extraction}
\label{alg:pipeline}
\begin{algorithmic}[1]
\REQUIRE CCTA $X$; coronary model $f_\theta$;
post-processing $\mathcal{P}$; anatomical model $g$
\ENSURE Coronary mask $M_C$; named trees $\mathcal{T}$;
PCAT measurements; EAT measurements; flags $\mathbf{q}$
\STATE $M_C\leftarrow\mathcal{P}(f_\theta(X))$
\STATE $\mathcal{A}\leftarrow g(X)$
\STATE $\mathcal{T}\leftarrow
       \textsc{BuildTree}(M_C,\mathcal{A})$
\STATE $\mathcal{T}\leftarrow
       \textsc{NameBranches}(\mathcal{T})$
\FOR{$v\in\mathcal{V}$}
    \STATE $C_v\leftarrow
           \textsc{SelectSegment}(\mathcal{T},v)$
    \STATE $(\mathrm{FAI}_v,V_v)\leftarrow
           \textsc{MeasurePCAT}(X,M_C,\mathcal{A},C_v)$
\ENDFOR
\STATE $(M_E,V_E)\leftarrow
       \textsc{RuleEAT}(X,\mathcal{A},\mathcal{T})$
\STATE $\mathbf{q}\leftarrow
       \textsc{QualityChecks}(\mathcal{T},\{C_v\},M_E)$
\IF{a non-contrast scan $X_{\mathrm{NC}}$ is available}
    \STATE $(K,A)\leftarrow
           \textsc{Calcium}(X_{\mathrm{NC}})$
\ENDIF
\end{algorithmic}
\end{algorithm}

\subsection{Coronary Tree Reconstruction and Branch Identification}
\label{sec:tree}

\subsubsection{Skeleton and graph representation}

The coronary mask is skeletonised using
three-dimensional thinning~\cite{lee1994skeleton}, producing the graph representation illustrated in Fig.~\ref{fig:anatomical_reasoning}(A).
The skeleton is represented as
\begin{equation}
G=(\mathcal{N},\mathcal{E}),
\qquad
r(\mathbf{p})=D_{M_C}(\mathbf{p}),
\label{eq:graph}
\end{equation}
where $\mathcal{N}$ contains endpoints and junctions,
$\mathcal{E}$ contains the centreline paths between them,
and $D_{M_C}$ is the Euclidean distance transform in mm.
The radius $r(\mathbf{p})$ is evaluated at centreline
point $\mathbf{p}$.

Each edge $e$ stores its centreline points, physical
length $\ell_e$, and mean radius $\bar r_e$.
This representation connects the predicted mask to the
geometry used for branch identification and PCAT sampling.

\subsubsection{Ostium detection and root assignment}

The aortic-root search region is
\begin{equation}
R=\{\mathbf{x}\in A_o:d(\mathbf{x},H)\leq12~\mathrm{mm}\}.
\label{eq:rootregion}
\end{equation}
Ostium candidates are connected components of the coronary
voxels near the root:
\[
\begin{aligned}
M_{\mathrm{near}}&=M_C\cap
\{\mathbf{x}:d(\mathbf{x},R)\leq2.5~\mathrm{mm}\},\\
\mathcal{C}_{\mathrm{ost}}&=\{C\in
\operatorname{CC}_{3D}(M_{\mathrm{near}}):|C|\geq30\}.
\end{aligned}
\]
Thus, proximity filtering precedes connected-component
identification and the minimum-size check.
For each candidate centre $\mathbf{c}$, its side is
determined relative to the local aortic-root centre
$\mathbf{a}$:
\begin{equation}
s(\mathbf{c})
=\frac{(c_x-a_x)+(c_y-a_y)}{\sqrt{2}}.
\label{eq:ostiumside}
\end{equation}
Coordinates follow the RAS convention: positive $x$ is
right, positive $y$ is anterior, and positive $z$ is
superior. Candidates with $s>0$ are assigned to the right
side; the remaining candidates are assigned to the left.

When several candidates occur on one side, the candidate
with the longest attached skeleton tree is retained.
Its root is the nearest skeleton node within 6~mm.
The ostium quality score is
\[
q_{\mathrm{ost}}
=\min\left(1,\frac{|s|}{10~\mathrm{mm}}\right).
\]

\subsubsection{Tree construction and quality assessment}

A tree is grown from each accepted root by breadth-first
traversal. Cycle-breaking and inter-tree bridge removal
follow the rules detailed in the Supplementary Information.
Spurs shorter than 2~mm are pruned, while components
unreachable from either root remain unassigned.
Automatic gap bridging is disabled; candidate bridges
are exported for reviewer inspection.

The heuristic edge score is
\begin{equation}
q_e=
\min\left(1,\frac{\ell_e}{10~\mathrm{mm}}\right)
\min\left(1,\frac{\bar r_e}{0.7~\mathrm{mm}}\right).
\label{eq:edgequality}
\end{equation}
Let $u=\ell_{\mathrm{unassigned}}/\ell_{\mathrm{total}}$
be the unassigned length fraction and $n_f$ the number
of fragments. The reconstruction status is
\begin{equation}
\mathrm{status}=
\begin{cases}
\text{complete candidate},
    &u\leq0.20\ \text{and}\ n_f\leq8,\\
\text{needs review},
    &\text{otherwise}.
\end{cases}
\label{eq:treestatus}
\end{equation}
This status describes the reconstruction criteria;
it does not establish anatomical correctness.

\subsubsection{Anatomical branch identification}

Names follow AHA/SCCT
nomenclature~\cite{austen1975,leipsic2014scct},
with assignments determined by geometric rules on trunk continuity, the anterior/posterior position of the LM children over their first 30~mm, and branch order (Supplementary Table~S15).
A child branch is considered substantial when its
subtree length is at least 10~mm. At a junction,
the trunk continues through the child with the
largest radius.

LM and RCA assignments receive a heuristic score of 0.9;
PDA and side-branch assignments receive 0.6.
For LAD and LCx,
\[
q_{\mathrm{branch}}
=\min\left(
0.9,\,
0.5+0.4\frac{|\Delta y|}{20~\mathrm{mm}}
\right),
\]
where $\Delta y$ is the anterior separation used by
the naming rule.
These scores are rule-based indicators rather than
calibrated probabilities, and branch names remain
available for expert confirmation.

\subsection{Pericoronary Adipose Tissue Quantification}
\label{sec:pcat}

\subsubsection{Segment and sampling-region definition}

PCAT sampling follows proximal coronary segment
definitions informed by prior
work~\cite{antonopoulos2017,oikonomou2018}.
The RCA segment extends from 10 to 50~mm from its
ostium; LAD and LCx segments extend from 0 to 40~mm
from the end of the LM.
Each selected centreline $C_v$ is resampled at 0.25~mm; Fig.~\ref{fig:anatomical_reasoning}(B) illustrates how its radius defines the sampling shell.

For voxel $\mathbf{x}$, let $\pi_v(\mathbf{x})$ be its
nearest point on $C_v$. The local sampling radius is
\begin{equation}
\begin{aligned}
\pi_v(\mathbf{x})
    &=\operatorname*{arg\,min}_{\mathbf{p}\in C_v}
      \|\mathbf{x}-\mathbf{p}\|_2,\\
r_v(\mathbf{x})
    &=\max\left\{
       r\big(\pi_v(\mathbf{x})\big),\,0.6~\mathrm{mm}
       \right\}.
\end{aligned}
\label{eq:pcatradius}
\end{equation}

Let $\Omega_0=\Omega\setminus(M_C\cup H\cup A_o)$
exclude coronary, heart, and aortic voxels.
With wall allowance $\delta=0.5$~mm, the sampling shell is
\begin{equation}
\begin{aligned}
S_v=\{\mathbf{x}\in\Omega_0:\;&
r_v(\mathbf{x})+\delta<d(\mathbf{x},C_v)\\
&\leq3r_v(\mathbf{x})+\delta\}.
\end{aligned}
\label{eq:shell}
\end{equation}
The inner boundary includes the wall allowance, and
the shell extends outward by twice the local radius.

\subsubsection{Fat measurements and reference comparisons}

The fat-intensity mask and vessel-specific PCAT mask are
\[
F_W=\{\mathbf{x}:-190\leq X(\mathbf{x})\leq-30\},
\qquad
F_v=S_v\cap F_W.
\]
PCAT attenuation and volume are then
\begin{equation}
\begin{aligned}
\mathrm{FAI}_v
    &=\frac{1}{|F_v|}
      \sum_{\mathbf{x}\in F_v}X(\mathbf{x}),\\
V_v&=\operatorname{Vol}(F_v).
\end{aligned}
\label{eq:fai}
\end{equation}
The mean attenuation is defined only for non-empty
$F_v$. A segment is flagged as short when less than
half of its nominal length is available.

Automatic-anatomy evaluation uses predicted masks,
centrelines, and branch identities; expert-anatomy
evaluation uses the corresponding expert references.
Both use the same measurement routine.
The paired-mask analysis in Fig.~\ref{fig:anatomical_reasoning}(D) holds scans, expert centrelines and measurement code fixed while replacing only the coronary mask source.

Here, FAI denotes uncorrected mean PCAT attenuation.
Exploratory threshold-crossing analyses and post hoc
sampling sensitivity analyses are described separately
in the evaluation protocol and Supplementary Information.

\subsection{Epicardial Adipose Tissue Without Pericardium Contours}
\label{sec:eat}

\subsubsection{Anatomical range and surrogate sac}

Figure~\ref{fig:anatomical_reasoning}(C) illustrates construction of the EAT compartment from the TotalSegmentator v2 fast anatomical masks~\cite{wasserthal2023}.
Its axial domain is
\[
\Omega_E=
\{\mathbf{x}:z_{\min}(H)\leq z(\mathbf{x})\leq z_{\mathrm{top}}\}.
\]
When an LM ostium is available,
\[
z_{\mathrm{top}}
=\min\{z_{\max}(H),z_{\mathrm{LM}}+35~\mathrm{mm}\};
\]
otherwise, $z_{\mathrm{top}}=z_{\max}(H)$.
The latter rule also applies to non-contrast scans
without a coronary tree.

Let $G_v$ contain the pulmonary veins and venae cavae.
The cardiac mass is
\begin{equation}
\begin{aligned}
G_{\mathrm{near}}&=\{\mathbf{x}\in A_o\cup G_v:
d(\mathbf{x},H)\leq8~\mathrm{mm}\},\\
M_{\mathrm{mass}}&=(H\cup G_{\mathrm{near}})
\setminus A_{\mathrm{desc}},
\end{aligned}
\label{eq:mass}
\end{equation}
where $A_{\mathrm{desc}}$ is the aortic region within
20~mm of the spine.

The surrogate sac is constructed in two steps:
\begin{equation}
\begin{aligned}
P_0&=\operatorname{hull}_z(M_{\mathrm{mass}})
     \oplus B_4,\\
P_1&=\{\mathbf{x}:d(\mathbf{x},M_{\mathrm{mass}})
                     \leq20~\mathrm{mm}\},\\
P&=(P_0\cap P_1\cap\Omega_E)\setminus O.
\end{aligned}
\label{eq:sac}
\end{equation}
Here, $\operatorname{hull}_z$ forms a convex hull on
each axial slice, $\oplus B_4$ denotes dilation by 4~mm,
and $O$ contains excluded lungs, bones, oesophagus,
trachea, and abdominal organs.
Thus, $P$ is a software-defined anatomical surrogate,
rather than an expert pericardium contour.

\subsubsection{Fat-component selection}

Fat candidates are $F_E=P\cap F_W$.
Only connected components lying within 3~mm of the
heart surface are retained:
\begin{equation}
M_E=
\bigcup_{\substack{
C\in\operatorname{CC}_{3D}(F_E)\\
d(C,\partial H)\leq3~\mathrm{mm}
}} C,
\label{eq:eat}
\end{equation}
where
$d(C,\partial H)=\min_{\mathbf{x}\in C}
d(\mathbf{x},\partial H)$.
The EAT volume is $V_E=\operatorname{Vol}(M_E)$.
Quality flags describe the EAT-to-heart volume ratio,
component count, and axial extent.
Fat outside the surrogate sac is reported separately.

\subsubsection{Learning from anatomical pseudo-labels}
Rule-derived pairs $\mathcal{D}_E=\{(X_i,M^{\mathrm{pseudo}}_{E,i})\}_{i=1}^{N}$ train a separate nnU-Net predictor, following self-training with generated targets~\cite{xie2020noisystudent}. Targets use the automatic tree's LM landmark (heart-top fallback) and no expert input; agreement with them measures rule reproduction. The estimators are evaluated separately: external rule inference uses automatic anatomy with the heart-top fallback, whereas the learned predictor needs only the image.

\subsection{Companion Calcium Scoring on Non-Contrast CT}
\label{sec:cac}

\subsubsection{Calcium segmentation and spatial constraints}
A single-channel ResEnc-L nnU-Net processes $X_{\mathrm{NC}}$ cropped to the TotalSegmentator heart with a 20~mm margin. Each predicted lesion is snapped to the nearest component of $Q_{130}=\{\mathbf{x}:X_{\mathrm{NC}}(\mathbf{x})\geq130\}$ within a 400~mm$^3$ search volume. Filtering retains lesions inside the heart dilated by 10~mm and outside the aorta dilated by 3~mm. Training and candidate inputs are described in Sections~\ref{sec:setup} and~\ref{sec:ablations}.

\subsubsection{Agatston calculation}
On slice $k$, let $\mathcal{C}_k$ contain retained two-dimensional lesions of area $a_c\geq1$~mm$^2$, with peak attenuation $I_c=\max_{\mathbf{x}\in c}X_{\mathrm{NC}}(\mathbf{x})$. For $I\geq130$~HU, density weight $w(I)=\min\{4,\lfloor I/100\rfloor\}$ gives the standard weights 1--4. The implemented slice-thickness-normalized Agatston score~\cite{agatston1990} is
\begin{equation}
A=\frac{t_z}{3~\mathrm{mm}}\sum_k\sum_{c\in\mathcal{C}_k}a_c\,w(I_c),
\label{eq:agatston}
\end{equation}
where $t_z$ is slice thickness. This module requires non-contrast CT.

\section{Experiments}
\label{sec:setup}
\label{sec:results}
\subsection{Datasets and Benchmark Construction}
Table~\ref{tab:data} summarizes the five datasets. ImageCAS uses four official 700/50/250 training/validation/test splits. Lumen and EAT models use the ImageCAS-X 560/80/160 partition. ImageCAS-X shares scans with ImageCAS, including 107 test cases present in the first ImageCAS training split; independently trained models therefore serve the two benchmarks. Across 800 paired references, whole-vessel masks have 3.4 times the lumen volume and Dice overlap 0.42, motivating separate evaluation of the annotation targets.

\begin{table*}[!t]
\caption{Datasets and evaluation tasks.}
\label{tab:data}
\centering\footnotesize
\begin{tabular}{@{}p{0.13\textwidth}p{0.25\textwidth}p{0.26\textwidth}p{0.31\textwidth}@{}}
\toprule
Dataset & Cohort and partition & Reference information & Evaluation role \\
\midrule
ImageCAS~\cite{zeng2023imagecas} & 1000 CCTA; four official 700/50/250 splits & Whole-vessel coronary masks & Whole-vessel segmentation; out-of-sample tree analysis \\
ImageCAS-X~\cite{imagecasx2026} & 800 shared CCTA; 560/80/160 & Lumen masks, 14 segments and centrelines & Lumen segmentation, branch names, PCAT references; EAT targets \\
COCA~\cite{coca2021} & Gated 582/78/116; non-gated 210/214 evaluable & Calcium masks and Agatston scores & Separate non-contrast calcium module \\
Visual Lab~\cite{rodrigues2016} & 20 non-contrast CT; tune-10/test-10 & Manual epicardial and mediastinal fat labels & External EAT-rule comparison \\
MultiD4CAD~\cite{prinzi2025multid4cad} & 118 CCTA; 108 usable EAT pairs & Radiologist EAT and LAD pericoronary masks; CAD label & Frozen-model external PCAT/EAT evaluation and tree QC \\
\bottomrule
\end{tabular}
\end{table*}

COCA is split by patient; two unreadable series and two missing heart masks leave 210 non-gated scans. Visual Lab uses tune-10/test-10 patients. MultiD4CAD contains 118 CCTA scans acquired at 120~kVp on one Siemens scanner. Stored intensities require HU-offset and axis/header correction. Nine empty EAT references and one incomplete payload leave 108 pairs; two missing heart labels overlap with these exclusions. LAD references are reformatted sub-volumes evaluated at patient level. ImageCAS NIfTI files do not provide tube voltage.

\subsection{Evaluation Metrics}
\label{sec:protocol}
\subsubsection{Segmentation accuracy and aggregation}
Let $P_i$ and $R_i$ be the predicted and reference masks for patient $i$. Dice measures spatial overlap~\cite{dice1945}:
\begin{equation}
D_i=\frac{2|P_i\cap R_i|}{|P_i|+|R_i|},\qquad
\overline D=\frac{1}{N}\sum_{i=1}^{N}D_i.
\label{eq:metric_dice}
\end{equation}
For the ImageCAS results, both configurations are evaluated across all four official test partitions. Because the partitions are equally sized, the unweighted mean of the partition means equals the patient-wise mean. ImageCAS-X is evaluated on 160 held-out patients. Validation and held-out test results are presented in separate tables. Dice and clDice scores are reported as percentages in the comparison tables, and all metrics are computed on the corresponding native image grids.

\subsubsection{Centreline, surface and connectivity metrics}
With $S(\cdot)$ denoting skeletonisation, clDice combines skeleton precision and sensitivity~\cite{shit2021cldice}:
\begin{equation}
\begin{aligned}
T_p&=\frac{|S(P)\cap R|}{|S(P)|},\quad
T_s=\frac{|S(R)\cap P|}{|S(R)|},\\
\mathrm{clDice}&=\frac{2T_pT_s}{T_p+T_s}.
\end{aligned}
\label{eq:metric_cldice}
\end{equation}
For surfaces $\partial P$ and $\partial R$, define the directed distance collection $\mathcal{D}_{P\to R}=\{d(\mathbf{x},\partial R):\mathbf{x}\in\partial P\}$. Following the ImageCAS-X evaluator~\cite{imagecasx2026}, symmetric HD95 is
\begin{equation}
\mathrm{HD95}=\max\{Q_{0.95}(\mathcal{D}_{P\to R}),
Q_{0.95}(\mathcal{D}_{R\to P})\},
\label{eq:metric_hd95}
\end{equation}
where $Q_{0.95}$ is the 95th percentile and distances are in mm~\cite{huttenlocher1993,taha2015metrics}. Full HD instead uses the maximum directed distances; it is distinguished from HD95 in Supplementary Table~S13. Connectivity discrepancy is $e_{\beta_0}=|\beta_0(P)-\beta_0(R)|$, where $\beta_0$ counts connected components. Published component counts alone are not substituted for this error. Coverage@1~mm measures the fraction of expert centreline covered within the 1~mm matching tolerance. Dice, clDice and coverage are higher-is-better; HD, HD95 and $e_{\beta_0}$ are lower-is-better. Equal component counts do not establish correct branch connectivity.

\subsubsection{Branch identification}
A predicted edge is correctly named when its assigned name matches the majority expert segment label along that edge. Per-edge precision is the proportion of predicted named edges satisfying this condition. A main vessel is correct in a scan if at least one of its predicted edges is correctly named. An expert segment is found when at least half of its centreline lies within 1~mm of a correctly named edge. The all-four rate requires correct RCA, LM, LAD and LCx identification in the same scan. These endpoints have different denominators and are reported separately.

\subsubsection{PCAT and EAT measurement agreement}
For paired automatic and reference measurements $a_i$ and $b_i$, define $\delta_i=a_i-b_i$. Bias and Bland--Altman limits of agreement (LoA) are~\cite{bland1986}
\begin{equation}
\mathrm{bias}=\overline\delta,\qquad
\mathrm{LoA}=\overline\delta\pm1.96s_\delta,
\label{eq:metric_agreement}
\end{equation}
where $s_\delta$ is the sample SD of paired differences. Bias is expressed in HU for attenuation and mL for volume; zero is ideal. Pearson $r$ describes association, while absolute-agreement ICC(2,1) assesses agreement between individual measurements~\cite{koo2016icc}. High correlation alone does not imply low bias. PCAT comparisons use available pairs within each artery. EAT overlap and volume comparisons use the reference slice range; additional fat-window matching is reported as a separate analysis, with its own reference definition.

\subsubsection{Calcium endpoints and uncertainty}
Calcium 3D Dice uses the 65 reference-mask-positive patients in the 116-patient gated test set. Slice Dice is first averaged within each patient's calcium-containing slices. Lesion sensitivity and positive predictive value are $TP/(TP+FN)$ and $TP/(TP+FP)$. Agatston agreement uses all 116 patients, with linearly weighted kappa~\cite{cohen1968kappa} for categories 0, $(0,100]$, $(100,400]$ and $>400$. Sensitivity and specificity at score thresholds are separate patient-level endpoints.

Available confidence intervals use patient-level percentile bootstrap resampling~\cite{efron1979}: 10\,000 resamples for the documented segmentation comparisons and PCAT/EAT analyses, and 2\,000 for calcium. Paired measurements are resampled together. Post hoc shell differences use paired Student-$t$ intervals. Uncertainty estimates apply to the corresponding model and evaluation population. The supplementary results retain 95\% percentile-bootstrap intervals for the documented ImageCAS-X comparisons and, where two were scored on the same scans, the paired bootstrap interval of the mean difference and a two-sided Wilcoxon signed-rank test; published comparisons are descriptive because comparator per-case outputs are unavailable. Exploratory CAD analyses use AUC and adjusted logistic regression; these are distinct from segmentation and agreement endpoints~\cite{maierhein2024metrics}.

\subsection{Implementation and Evaluation Protocol}
ResEnc-L uses the nnU-Net 40~GB plan with $160\times320\times256$ patches at $0.5\times0.35\times0.35$~mm. CT intensities are clipped to training-derived foreground percentiles and z-scored. Models train for 250 epochs with Dice plus cross-entropy and deep supervision; mirroring is disabled. Predictions return to the native grid. Benchmark sliding-window step is 0.25; whole-vessel out-of-sample and the archived MultiD4CAD inference use 0.5. The EAT predictor uses a low-resolution plan at $1.0\times0.69\times0.69$~mm.

Both ImageCAS configurations are evaluated across the four official splits, each comprising 700 training, 50 validation, and 250 test scans. The ensemble combines a baseline U-Net continued to 500 epochs, ResEnc-L, ResEnc-L fine-tuned for 250 epochs, and ResEnc-L trained for 1,000 epochs. The models are equally weighted, with a step size of 0.25 and a threshold of 0.5. Archived Split 1 validation experiments are reported separately from the updated summaries. For ImageCAS-X, M1 denotes the ResEnc-L lumen model, and M3 denotes the standard nnU-Net validation-ensemble member. Test-160 evaluates M1 using the validation-selected hysteresis settings and a 10 mm³ component filter. Downstream models retain their original provenance; MultiD4CAD uses frozen M1, the EAT network, and pretrained TotalSegmentator.

Training was performed on an NVIDIA A100 80 GB PCIe GPU using RunPod, Ubuntu 24.04, PyTorch 2.8.0, CUDA 12.8.1, and nnU-Net v2.8.1. Analysis used an RTX 4070 12 GB GPU and CPU, with Python 3.12.10, PyTorch 2.14.0/CUDA 12.6, and nnU-Net v2.8.1. TotalSegmentator 2.18.0 was run with the fast 3 mm total task. Stage throughput, test-set access, and protocol amendments are documented in the Supplement.

Published comparators were quoted without retraining. Table~\ref{tab:imagecas} lists published ImageCAS results and their partitions alongside the updated four-split summaries; Supplementary Table~S13 provides protocol details and distance metrics. Comparator scores in Table~\ref{tab:imagecasx} follow the ImageCAS-X Test-160 benchmark~\cite{imagecasx2026}. The Supplement reports available uncertainty estimates and paired tests for ImageCAS-X; comparisons with published point estimates are descriptive. Visual Lab includes repeated test configurations.

\subsection{Coronary Segmentation Results}
\begin{table*}[!t]
\caption{Comparison of whole-vessel coronary segmentation methods on ImageCAS.}
\label{tab:imagecas}
\centering\small
\renewcommand{\arraystretch}{1.12}
\begin{tabular*}{\textwidth}{@{\extracolsep{\fill}}llccc@{}}
\toprule
Method & Protocol & Dice (\%) $\uparrow$ & clDice (\%) $\uparrow$ & $e_{\beta_0}$ $\downarrow$ \\
\midrule
ImageCAS baseline~\cite{zeng2023imagecas} & Official 4 splits & 82.96 & -- & --  \\
Seg2RefineNet~\cite{seg2refinenet2025} & Official 4 splits & 83.13 & -- & --  \\
DiffCAS~\cite{diffcas2024} & -- & 84.59 & -- & --  \\
SADiff~\cite{xu2025sadiff} & 7:1:2 & 83.48 & -- & --  \\
MDSVM-UNet~\cite{yuan2026mdsvm} & 750/250 & 83.65 & -- & --  \\
STU-Net-L + vesselness~\cite{owusu2026bifurcation} & 750/250 & 81.80 & 89.00 & --  \\
STU-Net-L + softBCS~\cite{owusu2026bifurcation} & 750/250 & 81.30 & 87.90 & --  \\
STU-Net-L + SkeletonRecall~\cite{owusu2026bifurcation} & 750/250 & 82.00 & 87.40 & --  \\
nnU-Net + clDice~\cite{velasquez2026surface} & 800/200 & 81.64 & 86.78 & 5.14  \\
nnU-Net + cbDice~\cite{velasquez2026surface} & 800/200 & 81.92 & 86.99 & 4.53  \\
nnU-Net + surface matching~\cite{velasquez2026surface} & 800/200 & 82.05 & 87.12 & 4.05  \\
SwinUNETR + surface matching~\cite{velasquez2026surface} & 800/200 & 80.93 & 86.23 & 11.25  \\
NexToU + surface matching~\cite{velasquez2026surface} & 800/200 & 81.94 & 87.52 & 4.35  \\
\midrule
\textbf{Ours: single ResEnc-L} & Official 4 splits & 84.16 & 89.24 & 2.89 \\
\textbf{Ours: four-member ensemble} & Official 4 splits & \textbf{85.18} & \textbf{90.02} & \textbf{2.52} \\
\bottomrule
\end{tabular*}
\end{table*}

\begin{table*}[!t]
\caption{Comparison of coronary lumen segmentation methods on ImageCAS-X test-160.}
\label{tab:imagecasx}
\centering\small
\renewcommand{\arraystretch}{1.12}
\begin{tabular*}{\textwidth}{@{\extracolsep{\fill}}lcccc@{}}
\toprule
& \multicolumn{2}{c}{Overlap and centreline agreement} & \multicolumn{2}{c}{Surface and connectivity errors} \\
\cmidrule(lr){2-3}\cmidrule(lr){4-5}
Method & Dice (\%) $\uparrow$ & clDice (\%) $\uparrow$ & HD95 (mm) $\downarrow$ & $e_{\beta_0}$ $\downarrow$ \\
\midrule
TotalSegmentator~\cite{wasserthal2023} & 70.50 & 76.00 & 19.59 & 4.60 \\
3D-FFR-UNet~\cite{song2022ffr} & 84.90 & 89.80 & 15.93 & 4.80 \\
ADE-HTL~\cite{zhang2024adehtl} & 87.70 & 93.20 & 2.97 & 1.50 \\
Swin-UNETR~\cite{hatamizadeh2022swinunetr} & 87.90 & 92.50 & 3.18 & 3.80 \\
ImageCAS baseline~\cite{zeng2023imagecas} & 87.90 & 91.70 & 4.45 & 4.70 \\
nnU-Net~\cite{isensee2021} & 89.80 & 92.30 & 7.08 & 5.60 \\
nnU-Net + clDice~\cite{isensee2021,shit2021cldice} & 90.00 & 91.70 & 9.70 & 8.00 \\
CAS-Net~\cite{dong2023casnet} & 91.20 & 93.30 & 2.99 & 1.90 \\
\midrule
\textbf{Ours: M1 lumen model} & 91.01 & 93.90 & \textbf{2.39} & 3.26 \\
\textbf{Ours: M1 + post-processing} & \textbf{91.33} & \textbf{94.02} & 2.47 & \textbf{0.96} \\
\midrule
\multicolumn{5}{@{}l}{\textit{Human reference}} \\
Inter-observer agreement~\cite{imagecasx2026} & 92.80 & 95.40 & 2.46 & 0.40 \\
\bottomrule
\end{tabular*}
\end{table*}

In the updated ImageCAS summary, the single ResEnc-L model and four-member ensemble achieve Dice scores of 84.16\% and 85.18\%, respectively, across the four official splits (Table~\ref{tab:imagecas}). The ensemble improves on the single model by 1.02 percentage points and exceeds the published DiffCAS score by 0.59 points. Secondary metrics in Supplementary Table~S13 show higher clDice and lower HD95 for the ensemble, with a smaller change in component-count error. These comparisons are descriptive because the published studies use different evaluation partitions. Supplementary Table~S7 distinguishes the updated summaries from historical experiments.

On ImageCAS-X test-160, M1 and M1 with post-processing achieved Dice 91.01\% and 91.33\% (Table~\ref{tab:imagecasx}). Post-processing raised Dice by 0.32 percentage points (Wilcoxon $p<0.001$; 156 of 160 scans improved) and clDice by 0.12 points ($p=0.003$). Component-count error fell from 3.26 to 0.96, a 70.5\% reduction, while HD95 rose by 0.08~mm ($p=0.055$). Fragment removal therefore improved overlap and connectivity without improving the surface-distance endpoint. The post-processed model exceeds the published CAS-Net Dice by 0.13 points and has the highest clDice and lowest component-count error among the listed automatic methods; M1 alone has the lowest HD95. These comparisons describe published point estimates, without establishing statistical superiority over CAS-Net.

\subsection{Coronary Tree and Branch Identification Results}
\begin{figure*}[!t]
\centering
\includegraphics[width=0.60\textwidth]{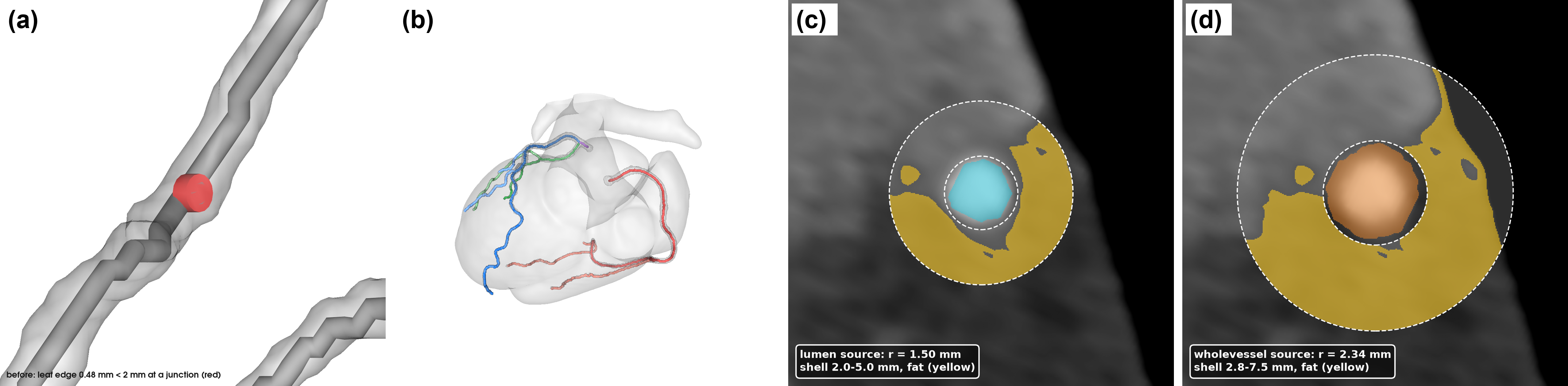}
\caption{Qualitative examples from a public ImageCAS-X scan. (a) Tree clean-up: a 0.48~mm leaf edge at a junction (red) is marked for pruning. (b) Named tree on the heart (RCA red, LM violet, LAD blue, LCx green). (c, d) The PCAT shell at the same RCA expert centreline point, 30~mm from the ostium, from the expert lumen mask (c; radius 1.50~mm, shell 2.0--5.0~mm) and the expert whole-vessel mask (d; radius 2.34~mm, shell 2.8--7.5~mm); fat voxels in yellow.}
\label{fig:qualitative}
\end{figure*}

Whole-vessel out-of-sample predictions cover all 1000 ImageCAS scans; 800 have expert segment references. Both ostia were detected in 799/800, and all four main vessels were correctly named in 697/800 (87.1\%; Table~\ref{tab:components_summary}A). All-four correctness was 89.5\% among quality-control (QC) complete candidates and 51.0\% among flagged cases. The original held-out lumen M1 outputs achieved 83.1\%. Side-branch edge precision ranged from 29\% to 57\%, confirming that geometric coverage does not establish naming accuracy. The gap between near-complete ostium detection and lower branch correctness lies in tree reconstruction and labeling; lower correctness among flagged scans supports their use as review candidates, although review-time savings were not measured (Fig.~\ref{fig:qualitative}).

\begin{table*}[!t]
\caption{Coronary branch identification, PCAT and EAT quantification, and calcium scoring results.}
\label{tab:components_summary}
\centering\small
\renewcommand{\arraystretch}{1.13}
\begin{tabular*}{\textwidth}{@{\extracolsep{\fill}}L{0.29\textwidth}cL{0.25\textwidth}L{0.34\textwidth}@{}}
\toprule
Evaluation & $n$ & Main endpoint & Supporting endpoint \\
\midrule
\multicolumn{4}{@{}l}{\textbf{A. Coronary branch identification}} \\
Whole-vessel, out-of-sample & 800 & All four correct: 87.1\% & Both ostia found: 799/800 \\
Lumen M1, test-160 & 160 & All four correct: 83.1\% & QC-complete candidates: 149/160 \\
\midrule
\multicolumn{4}{@{}l}{\textbf{B. PCAT: automatic versus expert anatomy (lumen M1, test-160)}} \\
RCA & 158 & Attenuation ICC: 0.97 & Bias: $+0.7$~HU; LoA [$-2.6$, $+4.1$] \\
LAD & 154 & Attenuation ICC: 0.93 & Bias: $+0.2$~HU; LoA [$-5.3$, $+5.7$] \\
LCx & 154 & Attenuation ICC: 0.95 & Bias: $-0.4$~HU; LoA [$-4.9$, $+4.1$] \\
\midrule
\multicolumn{4}{@{}l}{\textbf{C. EAT: external expert references}} \\
Visual Lab, rule & 10 & Dice: 66.4\% & Original manual labels \\
MultiD4CAD, rule & 108 & Dice: \textbf{72.5}\% & Volume $r$: \textbf{0.944}; bias: $-78$~mL \\
MultiD4CAD, learned & 108 & Dice: 70.8\% & Volume $r$: 0.626; bias: $\mathbf{-47}$~mL \\
MultiD4CAD, rule; matched window & 108 & Dice: 80.0\% & Slice range and fat window matched \\
\midrule
\multicolumn{4}{@{}l}{\textbf{D. Calcium: separate non-contrast COCA scans}} \\
Gated CT & 116 & 3D Dice: 88.1\% ($n=65$) & Weighted $\kappa$: 0.854; score ICC: 0.946 \\
Non-gated CT & 210 & -- & Weighted $\kappa$: 0.573; score ICC: 0.820 \\
\bottomrule
\end{tabular*}
\end{table*}

\subsection{Pericoronary Adipose Tissue Results}
\begin{figure*}[!t]
\centering
\includegraphics[width=\textwidth]{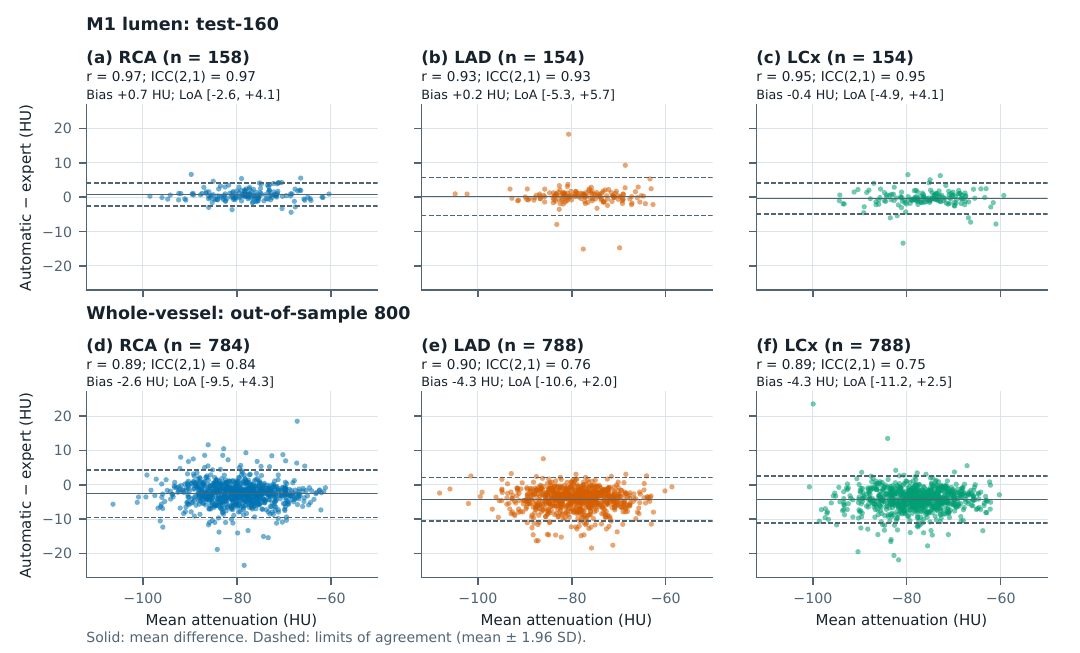}
\caption{Bland--Altman analysis of PCAT attenuation from automatic and expert coronary anatomy. (a--c) Lumen M1 on ImageCAS-X test-160. (d--f) Whole-vessel models on 800 out-of-sample scans. Columns show RCA, LAD and LCx; solid and dashed lines indicate mean bias and 95\% limits of agreement.}
\label{fig:pcat_agreement}
\end{figure*}

\textit{Automatic-anatomy agreement.} Original M1 outputs supplied 158 RCA, 154 LAD and 154 LCx measurements from 160 test scans. Figure~\ref{fig:pcat_agreement}(a--c) shows mean biases of $+0.7$, $+0.2$ and $-0.4$~HU, with attenuation ICCs of 0.97, 0.93 and 0.95 (Table~\ref{tab:components_summary}B). The limits of agreement remained within $-5.3$ to $+5.7$~HU. The plots compare one sampling routine on automatic and expert anatomy, not independently drawn PCAT masks.

For the whole-vessel models, Fig.~\ref{fig:pcat_agreement}(d--f) includes 784 RCA, 788 LAD and 788 LCx measurements from 800 out-of-sample scans. Biases were $-2.6$, $-4.3$ and $-4.3$~HU, with lower ICCs of 0.84, 0.76 and 0.75 and wider agreement limits. Strong attenuation correlation can thus coexist with systematic bias; the rows differ in model, population and inference step, so their contrast does not isolate annotation convention.

\textit{Annotation-source effects.} Using the paired replacement in Fig.~\ref{fig:anatomical_reasoning}(D), we held scans, expert centrelines, exclusions and code fixed in an 800-scan cohort. The primary analysis included 798 valid RCA pairs and 800 pairs each for LAD and LCx; whole-vessel references produced more negative attenuation than lumen references by 3.04, 4.24 and 3.78~HU for RCA, LAD and LCx. In an inclusive sensitivity analysis retaining the two flagged RCA pairs, the differences persisted under alternative shells (larger without the wall allowance, smaller with a fixed exterior-distance shell) and under a 1.0~mm radius floor (2.7--3.4~HU; Supplementary Table~S17). The control retains annotation-source differences alongside boundary convention; these multi-HU shifts under fixed scans and centrelines show why reference selection must accompany a reported PCAT value.

\textit{External PCAT.} MultiD4CAD yielded 90/118 paired LAD measurements with radiologist attenuation correlation $r=0.82$. Bias changed from $-2.6$ to $+3.6$~HU after reference fat-window restriction; corresponding LoA were [$-14.0$, $+8.9$] and [$-6.7$, $+13.8$]~HU. The bias reversal shows that reference definition can change the direction of apparent error; correlation alone does not establish interchangeability, and the unspecified reference extent prevents volume comparison.

\begin{figure*}[!t]
\centering
\includegraphics[width=0.94\textwidth]{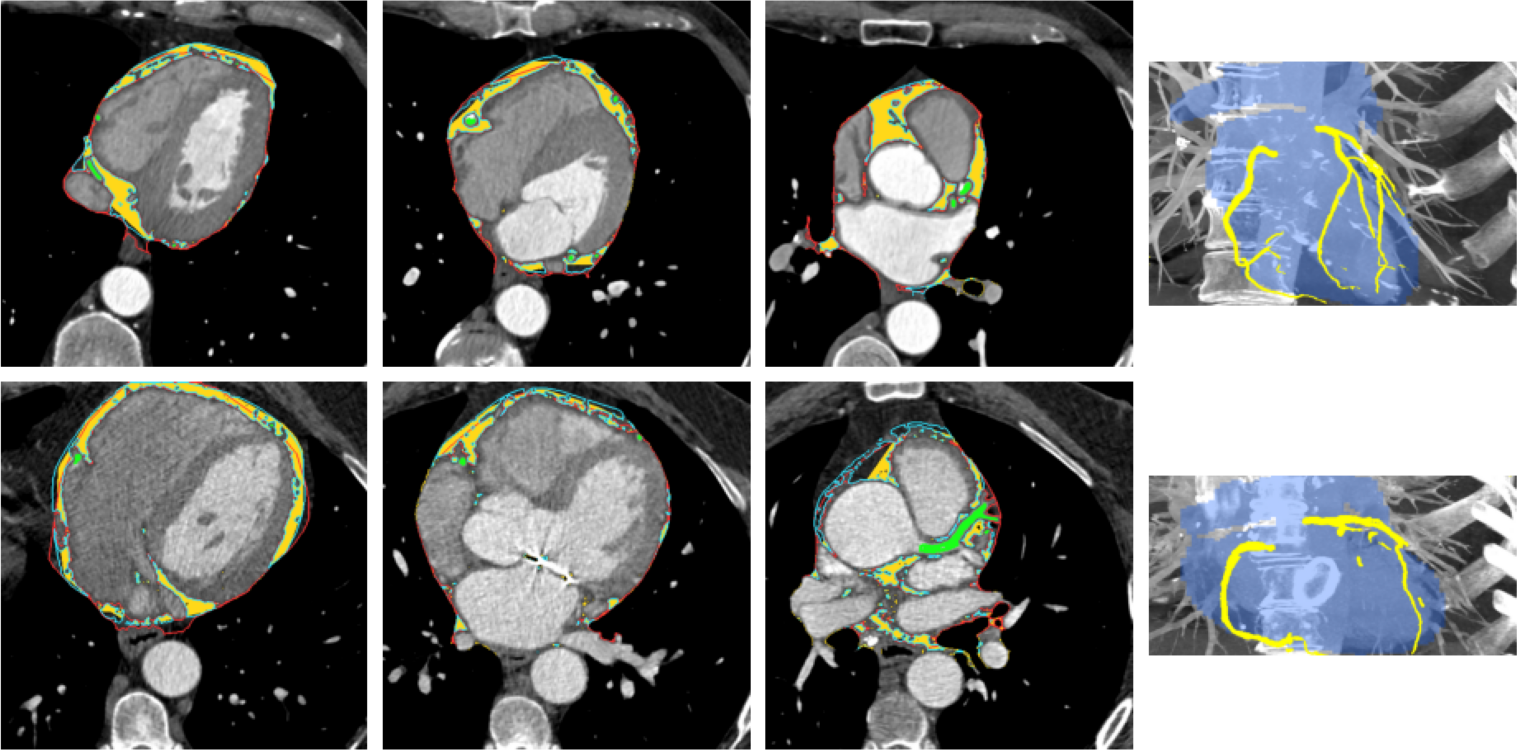}
\caption{EAT on MultiD4CAD: rule estimate (yellow), radiologist mask (red contour), self-trained model (cyan contour; expert-landmark version, Supplementary Table~S19) and lumen prediction (green) on two patients, with the coronal lumen projection.}
\label{fig:eat_examples}
\end{figure*}

\subsection{Epicardial Adipose Tissue Results}
On Visual Lab test-10, rule-based EAT achieved Dice 0.664, increasing to 0.705 when pixels under the drawn pericardium line were ignored. Sensitivity 0.83 and PPV 0.57 indicated over-segmentation. The learned predictor (retrained with automatic LM landmarks; Supplementary Table~S19) reproduced its held-out rule targets with Dice 0.856 and volume correlation 0.997. This establishes target reproduction, not agreement with an independently defined pericardium.

On 108 MultiD4CAD pairs, the rule achieved Dice 0.725 and volume correlation 0.944 within the reference slice range, compared with 0.708 and 0.626 for the learned predictor (Table~\ref{tab:components_summary}C). Fat-window matching raised rule Dice to 0.800, a separate reference-conditioned endpoint. Rule volume bias remained $-78$~mL with LoA [$-203$, $+47$]~mL; the learned predictor had lower absolute bias ($-47$~mL) but weaker correlation. Automatic instead of expert landmarks changed the targets in 84 of 800 scans and MultiD4CAD Dice by $-0.007$ (95\% CI $-0.010$ to $-0.005$; Supplementary Table~S19). Post hoc, the attachment distance (2--4~mm) changed rule Dice by less than 0.001 and bias by at most 0.2~mL, and the superior limit (LM + 25--45~mm or heart top alone) by at most 0.010 and 4.4~mL (Supplementary Tables~S16 and S18). Figure~\ref{fig:eat_examples} shows two patients; Supplementary Fig.~S6 shows volume agreement and overlap distributions. In this archived rule evaluation, the LM landmark was found in 98 of 108 scans, with heart-top fallback otherwise. About 16\% of reference voxels lay outside the fat window and sac coverage was 75\%, so reference and estimation differences both contribute. The learned model's smaller bias but wider limits (Supplementary Fig.~S6b) show that average error and patient-level reliability need separate assessment.

\subsection{Companion Calcium Scoring Results}
On gated COCA, 3D Dice was 0.881 across 65 reference-mask-positive scans, with lesion sensitivity 0.903 and PPV 0.873 (Table~\ref{tab:components_summary}D). Across all 116 patients, weighted kappa was 0.854 and ICC 0.946; score bias was $+34$ with LoA [$-322$, $+391$]. On 210 non-gated scans, kappa decreased to 0.573 and ICC to 0.820. Pacemaker leads and aortic or valve leakage caused marked errors. Because Agatston scoring weights lesion area by density, small high-attenuation false positives can affect scores disproportionately. The gated/non-gated gap consequently motivates acquisition-specific score validation. These results apply to non-contrast CT; detailed intervals are retained in the supplement.

\subsection{Ablation and Configuration Selection}
\label{sec:ablations}
On validation-80, hysteresis and component filtering raised M1 Dice from 91.32\% to 91.61\% and reduced component-count error from 3.23 to 0.90 (Supplementary Table~S14). HD95 increased and coverage decreased slightly; ensembling M1 and M3 did not improve Dice. The test-160 result appears in Table~\ref{tab:imagecasx}.

Single-run validation ablations favored ResEnc-L without mirroring: for heart-crop calcium, ResEnc-L added 21.8 Dice percentage points and mirroring removed 5.0 points. On Split~1 validation, ResEnc-L added 2.1 points and mirroring removed 3.5 points relative to the archived baseline U-Net (Supplementary Fig.~S7). A 1000-epoch ResEnc-L alone reduced Dice by 0.19 points; the four-member ensemble achieved Dice 0.8469 in that archived Split~1 validation experiment. Skeleton Recall reduced Dice by 0.22 points while increasing centreline coverage by 2.6 points. The validation experiments thus expose trade-offs between overlap and structural coverage rather than uniform gains across metrics.

\subsection{Whole-Framework External Evaluation}
\label{sec:external}
Frozen M1 inference at step 0.5 without post-processing on 118 MultiD4CAD scans produced a median of 13 components, versus five internally. Both ostia were found in 100/118 scans; QC-complete candidates decreased from 149/160 (93.1\%) internally to 37/118 (31.4\%) externally. These are internal QC designations.
Re-running the 118 scans with the configuration locked on ImageCAS-X validation (step 0.25, hysteresis, 10~mm$^3$ filter) halved the median component count to 6.5, found both ostia in 98/118 and raised QC-complete candidates to 54/118 (45.8\%). LAD PCAT was measurable in 91 scans ($r=0.78$, bias $-3.2$~HU); on the 83 scans measured in both runs, the mean attenuation change was $-0.2$~HU (SD 0.9), and correlation with the radiologist was similar (0.83 versus 0.84). With the rerun tree, the EAT rule detected the LM landmark in 101/108 scans; Dice (0.725) and volume bias ($-78$~mL) were unchanged at the reported precision.
External PCAT was available for 96, 90 and 90 vessels (RCA, LAD, LCx; 93, 91 and 91 after re-running), with 108 EAT pairs. Exploratory CAD associations were weak: EAT-volume AUCs were 0.40--0.44, and PCAT-attenuation AUCs were 0.44--0.57 in the original evaluation and 0.44--0.53 after re-running, with no significant adjusted odds ratios. These prevalent-CAD endpoints differ from the inflammatory and prognostic associations previously studied for PCAT~\cite{antonopoulos2017,oikonomou2018}. The small, single-scanner cohort limits interpretation of these exploratory associations.

\section{Discussion and Conclusion}
\label{sec:discussion}
\label{sec:conclusion}
\subsection{From Segmentation Accuracy to Anatomical Measurements}
Lumen Dice of 0.9133 and clDice of 0.940 support reconstruction, but overlap and anatomical usability remain distinct. The updated ImageCAS summary favors ensembling across overlap and surface measures, with a smaller connectivity gain; lumen post-processing removes fragments with a small HD95 penalty. These differences motivate multiple endpoints~\cite{maierhein2024metrics} and explicit branch and sampling-region reconstruction.

Compared with automated PCAT workflows~\cite{nguyen2024pcat}, this framework adds EAT estimation and training targets through shared anatomical construction (Fig.~\ref{fig:anatomical_reasoning}). Graph geometry locates vessel-specific regions, surrounding anatomy constrains EAT, and paired mask replacement exposes reference sensitivity.

\subsection{Reference Definitions and Measurement Interpretation}
Larger whole-vessel radii shift and widen the PCAT shell, partly explaining more negative attenuation. Residual differences under alternative shells and radius floors show that radius alone is insufficient. Mask replacement changes annotation source and boundary convention together; comparisons must therefore preserve reference definitions even when automatic-versus-expert agreement is high.

EAT results separate association from agreement: the rule better preserves volume ordering but systematically underestimates volume. The image-only learned predictor requires neither frame nor tree at inference and reduces absolute bias ($-47$ versus $-78$~mL), at the cost of weaker correlation. Its training targets use automatic landmarks. Fat-window matching improves overlap without repairing compartment geometry; accurate target reproduction does not establish expert-level pericardium recovery. Supervised non-contrast studies~\cite{commandeur2018} use different targets and cannot be ranked directly.

\subsection{External Generalization and Practical Implications}
The lower external complete-candidate rate exposes reconstruction failures beyond internal segmentation accuracy. Missing measurements and quality flags identify outputs needing inspection, although their triage value remains untested. The complete lumen configuration raised the rate from 31.4\% to 45.8\% while EAT agreement and common-case PCAT attenuation remained similar (Section~\ref{sec:external}). The remaining gap indicates reconstruction challenges beyond the tested inference settings. PCAT threshold transfer remains acquisition- and reference-dependent~\cite{ma2020fai,oikonomou2018}.

Low side-branch precision motivates stronger anatomical constraints or learned labeling~\cite{wu2019labeling}. Likewise, calcium overlap does not prevent score errors from high-density artifacts or leakage, particularly on non-gated CT~\cite{vanvelzen2020}. Task-specific validation is therefore necessary before the combined measurements can support clinical risk prediction.

\subsection{Limitations and Conclusion}

Several limitations should be considered when interpreting these results. Comparisons with published methods are descriptive because evaluation protocols and test partitions differ, and comparator per-case outputs were unavailable for matched statistical analysis. Therefore, the reported differences do not establish that one method is statistically superior to another. In addition, improved coronary segmentation did not, by itself, improve agreement for downstream PCAT or EAT measurements. These measurements retain the provenance and limitations of their original models, so their performance cannot be attributed to the segmentation model alone.

The EAT training targets were generated using a lumen model trained on the same training scans. This creates a potential dependence between the model used to construct the targets and the data used to train the downstream EAT model; the effect of this dependence on target quality and downstream performance has not been independently assessed. Several aspects of model and workflow performance also remain unevaluated, including full-output EAT Dice, patient-level latency, review-time savings, and the utility of quality flags. Sensitivity analyses for shell thickness, radius floor, attachment distance, and EAT limits were conducted post hoc, so their results should be regarded as exploratory rather than as evidence of a prospectively validated operating range.

The measurements may also be affected by anatomical variation and image acquisition conditions. Further evaluation across broader cohorts and imaging settings is needed to determine how consistently the methods perform and when expert review is required. Independent generation or assessment of EAT targets, a reader study, and outcome validation would help establish the reliability and clinical relevance of the measurements. Until these evaluations are completed, the framework should be viewed as a research approach rather than as evidence of established clinical utility.

\bibliographystyle{IEEEtran}
\bibliography{references}
\end{document}